\documentclass[runningheads,a4paper]{llncs}

\usepackage{amssymb}
\usepackage{graphicx}

\usepackage{color}

\usepackage{url}
\urldef{\mailsa}\path|m.moustafa@aucegypt.edu|
  
\newcommand{\keywords}[1]{\par\addvspace\baselineskip
\noindent\keywordname\enspace\ignorespaces#1}

\begin{document}

\mainmatter  

\def\PSIVT2015SubNumber{132} 

\title{Applying deep learning to classify pornographic images and videos}

\titlerunning{Applying deep learning to classify pornographic images and videos}

%
%
\author{Mohamed N. Moustafa}
\authorrunning{M. Moustafa}

\institute{Department of Computer Science and Engineering, \\
	The American University in Cairo,\\
New Cairo 11835, Egypt\\
\mailsa}

%
%

\toctitle{Applying deep learning to classify pornographic images and videos}
\tocauthor{Mohamed Moustafa}
\maketitle

\begin{abstract}
It is no secret that pornographic material is now a one-click-away from everyone, including children and minors. General social media networks are striving to isolate adult images and videos from normal ones. Intelligent image analysis methods can help to automatically detect and isolate questionable images in media. Unfortunately, these methods require vast experience to design the classifier including one or more of the popular computer vision feature descriptors. We propose to build a classifier based on one of the recently flourishing deep learning techniques. Convolutional neural networks contain many layers for both automatic features extraction and classification. The benefit is an easier system to build (no need for hand-crafting features and classifiers). Additionally, our experiments show that it is even more accurate than the state of the art methods on the most recent benchmark dataset.
\end{abstract}

\section{Introduction}

For one reason or another, we aim for filtering out adult images and videos. Most parents, for instance, strive to protect their young children and minors from accessing online porn web sites either intentionally or accidentally. Additionally, in educational and workplace settings, basic ethical and conduct standards dictate that such images become inaccessible to the community.
 
There are many indicators that online pornographic content is increasing exponentially \cite{pornsites}, especially in the past few years \cite{Short2012}.
Given the continuous flood of uploaded images and videos to general social media networks, one can imagine the difficult and tedious job of
filtering out inappropriate material manually by the sites administrators.
Therefore, we need to turn our attention to a solution that automatically detects and isolates porn content. Obviously, the solution is a form of an intelligent machine that can analyze the text, audio, or visual signals.

In this paper, we focus on the visual cue, being the most salient form of pornography.
We propose a deep learning system that automatically analyzes images (and video frames) before classifying the content as regular or porn. Our proposed solution achieved the highest classification rate, to our knowledge, of more than 94\%, on the most recent Pornography benchmark dataset \cite{NPDIdataset}.

\section{Related work}
Most of the existing porn detection research relies on image analysis, based on some form of machine intelligence. Usually, the intelligent processing pipeline includes a feature extractor followed by a classifier. The feature extractor abstracts important information from the input image and feeds it to the pre-trained classifier that takes the decision. Along with the features, mentioned below, Support Vector Machines (SVM) was the most frequently used classifier in recent methods. 

Historically, the most investigated feature was the human skin color \cite{jones1999}. If the input image contains too much skin colored pixels, it was taken as an indicator of nudity. However, skin color solely is not reliable since a face closeup image has a lot of skin pixels while being non-porn. So, researchers augmented the skin color with other constraints or shape features \cite{rowley2006}. With the introduction of new computer vision models, porn detection algorithms became more accurate. For instance, Deselaers et al introduced classification accuracy improvement by adopting the visual bag-of-words (BoW) model \cite{Deselaers2008}. BoW tries to extract the most common patches that exist on a set of training images. For a detailed survey of existing methods, please refer to \cite{Ries2014}.

Recently, Avila et al \cite{Avila2013453} introduced a BoW framework, the BossaNova representation, to classify pornographic videos. BossaNova relies on the HueSIFT descriptors. Since HueSIFT represents both color and shape, it outperformed standard BoW models that rely only on shape or edge cues. Additionally, Caetano et al \cite{Caetano2014} extended the BossaNova to be more suitable for video classification. They applied binary descriptors with BossaNova for even better accuracy on the same benchmark dataset.

\section{Deep learning}
In parallel with the traditional route of handcrafting a set of features and a classifier,  another route is witnessing a rising interest: \em deep learning\em. This route combines both features extraction and classification into one module with less involvement from the designer in terms of selecting features or classifier. The ultimate target is to have a generic architecture that can learn any problem from its data, thus coming closer in performance to the human brain. Many researchers are now paying more attention to deep learning systems, specifically after the Alex Krizhevsky et al (AlexNet) outstanding performance in the  ILSVRC-2012 Imagenet competition \cite{alexnet2012}. Imagenet challenge includes classifying hundreds of thousands of images to 1000 different possible classes. Krizhevsky used a convolutional neural network (ConvNet) architecture to win this most challenging competition in the area of image classification. Since then, ConvNet appeared in solutions to numerous image analysis problems, e.g., pedestrian detection, traffic sign recognition, image segmentation, image restoration, and object recognition. Recently, the winner of 2014 Imagenet competition was also a ConvNet based system: GoogLeNet \cite{googlenet2014}.

In this paper we propose applying  a system of ConvNets to solve the pornographic classification with higher accuracy than the reported state of the art performance \cite{Caetano2014}.

\section{Proposed method}
We propose applying a combination of ConvNets to classify porn from regular images and video frames. We will first describe the slight modifications we propose to change in the existing AlexNet and GoogLeNet to suit our problem. This is followed by a proposed simple fusion of both networks.

\subsection{ANet: AlexNet-based classifier}
\begin{figure}
	\centering
	\includegraphics[width=0.9\linewidth]{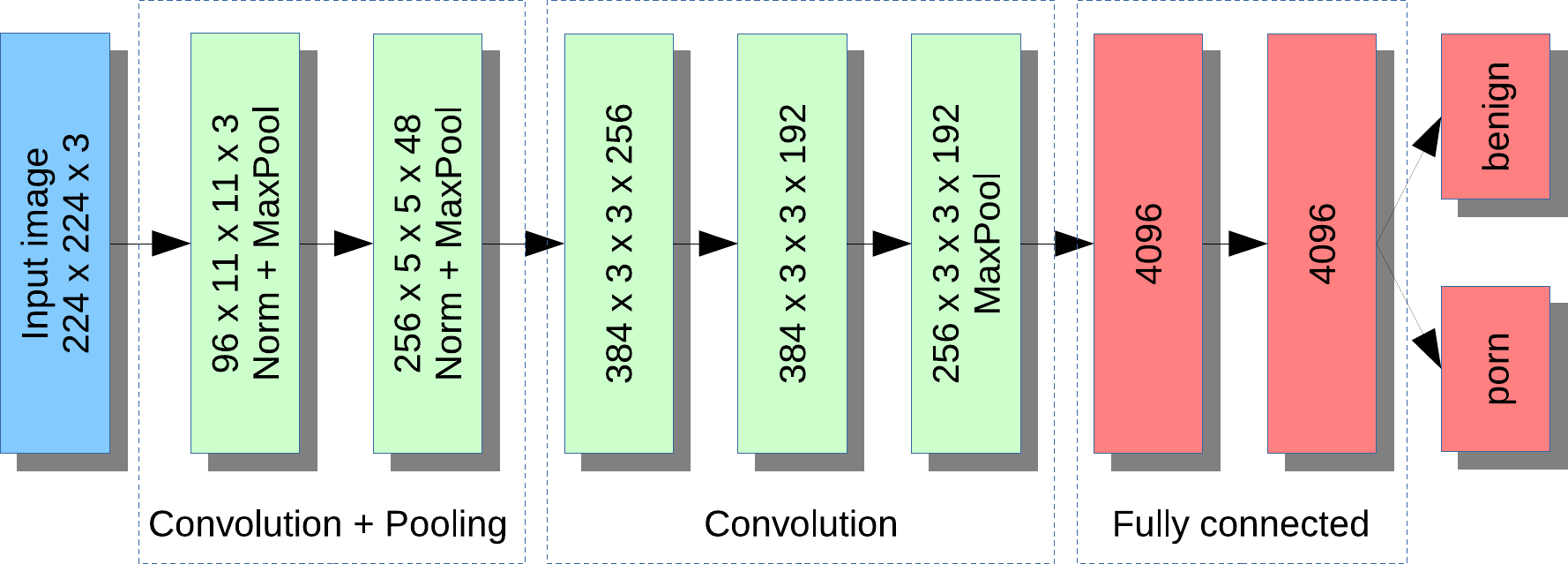}
	\caption{ANet: AlexNet-based deep ConvNet porn image classifier. The first five layers (in green) are convolutional followed by three fully connected layers (in red), including the output layer that contains two neurons corresponding to our two classes `benign' or `porn'.}
	\label{fig:AlexNet-classifier}
\end{figure}

Basically, we adopt the same architecture proposed in \cite{alexnet2012}, sans the output layer. As depicted in figure \ref{fig:AlexNet-classifier}, the net contains five convolutional layers of neurons followed by three fully-connected ones. Each convolution layer filters its input two dimensional vectors with a kernel, whose coefficients are calculated iteratively during the training process. The fully connected layers are simply implementing the dot product between the input and weight vectors, where each neuron in layer $l$ is connected to all outputs of neurons in layer $l-1$. All neurons have Rectified Linear Unit (ReLU) activation function to speed up the learning process. The output of the last fully-connected layer is fed to a 2-way softmax which produces a distribution over our two class labels: benign or porn.

ANet contains almost 56 million parameters that need to be calculated during the learning process. In order to properly train such a large
network, a huge dataset is needed. The collection and labelling of this
dataset would prove impractical.
 So, instead of starting training from scratch, we trained only the last layer that we newly introduced to the reference network. We started from the reference AlexNet weights for the first seven layers of the ConvNet. Those weights are the results of training the original AlexNet on the Imagenet 1.2 million images from 1000 different classes  \cite{jia2014caffe}.
During our training, we used benign and porn images as input to the first layer, computed the output of the seventh layer (the feature vector), and only changed the weights of the eighth layer in a supervised way given the ground truth label of the image. This way, we are using the ConvNet as a general feature extractor (the first seven layers) and as a classifier (the last layer). The last layer could also be replaced with any other classifier, e.g., Support Vector Machine (SVM). This fine-tune training method lets the network train faster with less numbers of training data than the full method as it has less parameters to adapt (those of the last layer only).

To test a new image, we adjust it by rescaling and subtracting the mean, before feeding it as input to the trained ConvNet. The network has two normalized outputs corresponding to the confidence in each class that sum to 100\%. The test image is labeled after the output neuron with the largest score.

\subsection{GNet: GoogLeNet-based classifier}
\begin{figure}
	\centering
	\includegraphics[width=0.9\linewidth]{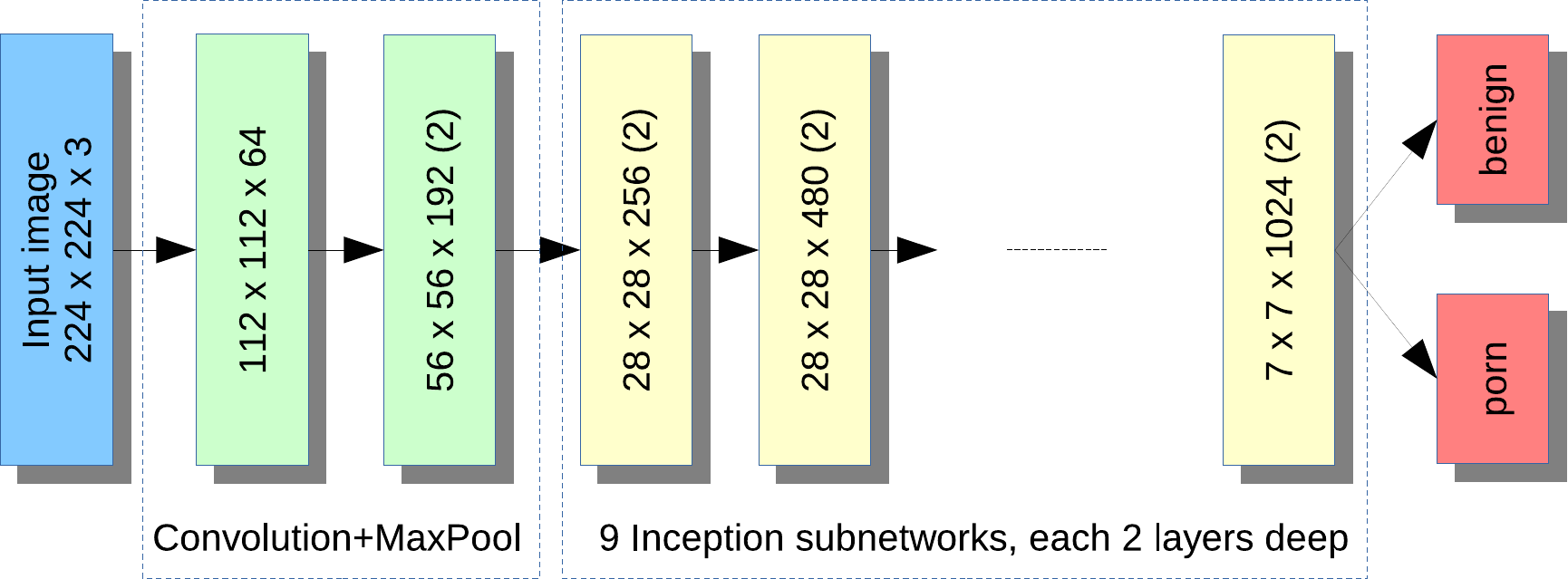}
	\caption{GoogLeNet-based deep ConvNet porn image classifier. The first three layers (in green) are convolutional followed by nine inception modules (in yellow), each contains two convolutional layers. The output layer is fully connected and contains two neurons corresponding to our two classes `benign' or `porn'.}
	\label{fig:GoogLeNet-classifier}
\end{figure}

GoogLeNet, shown in figure \ref{fig:GoogLeNet-classifier}, comprises 22 layers \cite{googlenet2014}. It is much deeper than AlexNet and incorporates the \em inception \em concept, which performs dimensionality reduction and projection. Each inception module contains two convolutional layers. Again, we have to change the output layers to produce only two labels: benign or porn. We follow the same steps in modifying, training, and testing this ConvNet as we have explained in the ANet classifier subsection above.

\subsection{AGNet: Ensemble-ConvNet classifier}
\begin{figure}
	\centering
	\includegraphics[width=0.9\linewidth]{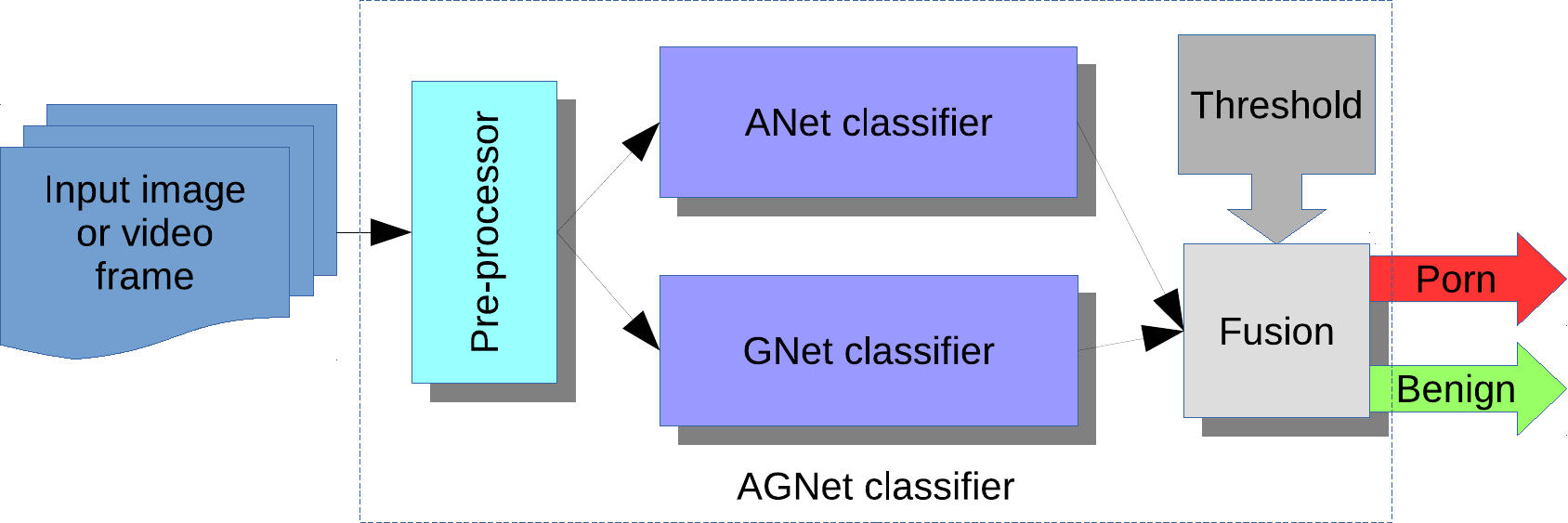}
	\caption{Proposed AGNet porn image classifier.}
	\label{fig:AGNet-classifier}
\end{figure}

When solving a hard problem, what would be better than consulting an expert? It would be having more than one to consult!
We expect that ANet and GNet are making different classification mistakes. Therefore, we believe that Our \em Ensemble-ConvNet \em, fusing both ANet and GNet scores, should result in a smoother decision that makes less mistakes. We describe in figure \ref{fig:AGNet-classifier} the overall architecture of our proposed AGNet classifier. We have followed the same procedure as \cite{alexnet2012} to pre-process images before feeding it as input to the first convolution layer. The preprocessing step in this case
includes rescaling the image to 256 x 256 RGB pixels, subtracting the mean image, and finally dividing it into one or more windows of 224 x 224 pixels each to increase the training data instances. Each of ANet and GNet tests the input image, and produces `benign' and `porn' scores. We combine those scores in the `Fusion' block, and finally decide whether it belongs to the benign or porn classes based on some predetermined threshold. By default, this threshold is 50\%. This fusion block could be as complex as another trained nonlinear classifier. On the other hand, it could be as simple as a linear weighted average of scores. We show in our experiments below that a simple weighted average with equal weights is enough to produce superior accuracy than each individual network.

To test a video sequence, we extract key frames and test them individually and finally label the video based on majority voting. In the event that the number of benign and porn voted frames is equal, we decide based on the class with the largest sum of scores for all key frames.

\section{Experiments and results}
We have implemented our proposed ANet, GNet, and AGNet classifiers in C++ relying on the open source frameworks CAFFE \cite{jia2014caffe} to train and test the ConvNets. To put our results in comparison with other state of the arts methods, we have used a recent benchmark dataset described below.

\subsection{NDPI dataset}
The NPDI Pornography database contains nearly 80 hours of 400 pornographic and 400 non-pornographic videos. It has been collected by the NPDI group, Federal University of Minas Gerais (UFMG), Brazil \cite{NPDIdataset}. The database consists of several genres of pornography and depicts actors of many ethnicities, including multi-ethnic ones. The NPDI group has collected the pornographic videos from websites which only host that kind of material. The non-pornographic videos include two subcategories: 200 videos chosen at random (which they called ``easy'') and 200 videos (``difficult''), selected from textual search queries like ``beach'', ``wrestling'', ``swimming'', which contain body skin but not nudity or porn. The videos are already segmented into shots. On the average, there are almost 20 shots per video. A key-frame has been selected as simply the middle-frame of each video shot. In total, there are 16,727 key-frames (shots). Tables \ref{table:NPDI content} and \ref{table:NPDI samples} highlight some statistics and samples of this dataset respectively. To unify the cross-validation protocol, the dataset is divided into five folds by generating nearly 640 videos for training and 160 for testing on each fold. In our experiments, we have followed the same training/testing 5-folds cross-validation.

\begin{table}
	\caption{NPDI dataset content}
	\centering
	\begin{tabular}{cccc}
		\hline \hline
		Class &	Videos & 	Hours & 	Shots per Video \\
		\hline 
		Porn &	400 & 	57 & 15.6 \\
		Non-porn (``easy'') &	200 & 11.5 &	33.8 \\
		Non-porn (``difficult'') & 	200 & 8.5 & 17.5 \\
		\hline 
		All videos & 800 & 77 & 20.6 \\
		\hline
	\end{tabular}
	\label{table:NPDI content}
\end{table}

\begin{table}
	\caption{NPDI dataset samples}
	\centering
	\begin{tabular}{c|c|c}
		\hline \hline
		Porn & Non-porn (``difficult'') & Non-porn (``easy'') \\ 
		\includegraphics[width=0.3\linewidth]{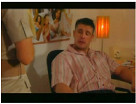} & \includegraphics[width=0.3\linewidth]{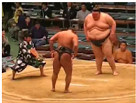} & \includegraphics[width=0.3\linewidth]{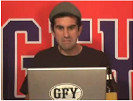} \\ 
		\includegraphics[width=0.3\linewidth]{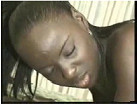} & \includegraphics[width=0.3\linewidth]{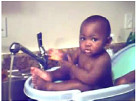} & \includegraphics[width=0.3\linewidth]{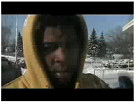} \\ 
		\hline 
	\end{tabular}
	\label{table:NPDI samples}
\end{table}


\subsection{Experiment: testing individual images}
In this experiment, we have tested each classifier on the NPDI videos key-frames as individual images. The objective of this experiment was threefold. First, it was a comparison between the full versus the fine-tune training of the ConvNets. Second, we wanted to check whether GoogLeNet was better than AlexNet in this problem as it was in the Imagenet categorization challenge. Third, we were testing the individual ConvNet classifier versus the ensemble one. We have tested many variations of the `fusion' block mentioned in figure \ref{fig:AGNet-classifier}. Our findings do not show much differences in the results.
Consequently, we decided to choose the equal weighted average of both ConvNets scores for simplicity. 

In total, we trained 5 ANets, 5 GNets using four of the NPDI five folds of videos and used the fifth one for testing. We recorded the average accuracy of classification for various values of the threshold from 0 to 100 to build Receiver Operating Characteristics (ROC) curves. ROC is very informative since the user can select his or her operating point based on the application. Simply put, at threshold 0, both correct benign and false porn rates will be 1.0, which corresponds to the top right corner point of the ROC. Increasing the threshold tightens the classifier filtration with lowering both rates. On the other hand, at threshold 100, the classifier will not pass any image and all classification rates will be 0.0, which is the left bottom point of the full ROC.

\begin{figure}
	\centering
\begingroup
\makeatletter
\providecommand\color[2][]{%
	\GenericError{(gnuplot) \space\space\space\@spaces}{%
		Package color not loaded in conjunction with
		terminal option `colourtext'%
	}{See the gnuplot documentation for explanation.%
}{Either use 'blacktext' in gnuplot or load the package
color.sty in LaTeX.}%
\renewcommand\color[2][]{}%
}%
\providecommand\includegraphics[2][]{%
	\GenericError{(gnuplot) \space\space\space\@spaces}{%
		Package graphicx or graphics not loaded%
	}{See the gnuplot documentation for explanation.%
}{The gnuplot epslatex terminal needs graphicx.sty or graphics.sty.}%
\renewcommand\includegraphics[2][]{}%
}%
\providecommand\rotatebox[2]{#2}%
\@ifundefined{ifGPcolor}{%
	\newif\ifGPcolor
	\GPcolortrue
}{}%
\@ifundefined{ifGPblacktext}{%
	\newif\ifGPblacktext
	\GPblacktextfalse
}{}%
\let\gplgaddtomacro\g@addto@macro
\gdef\gplbacktext{}%
\gdef\gplfronttext{}%
\makeatother
\ifGPblacktext
\def\colorrgb#1{}%
\def\colorgray#1{}%
\else
\ifGPcolor
\def\colorrgb#1{\color[rgb]{#1}}%
\def\colorgray#1{\color[gray]{#1}}%
\expandafter\def\csname LTw\endcsname{\color{white}}%
\expandafter\def\csname LTb\endcsname{\color{black}}%
\expandafter\def\csname LTa\endcsname{\color{black}}%
\expandafter\def\csname LT0\endcsname{\color[rgb]{1,0,0}}%
\expandafter\def\csname LT1\endcsname{\color[rgb]{0,1,0}}%
\expandafter\def\csname LT2\endcsname{\color[rgb]{0,0,1}}%
\expandafter\def\csname LT3\endcsname{\color[rgb]{1,0,1}}%
\expandafter\def\csname LT4\endcsname{\color[rgb]{0,1,1}}%
\expandafter\def\csname LT5\endcsname{\color[rgb]{1,1,0}}%
\expandafter\def\csname LT6\endcsname{\color[rgb]{0,0,0}}%
\expandafter\def\csname LT7\endcsname{\color[rgb]{1,0.3,0}}%
\expandafter\def\csname LT8\endcsname{\color[rgb]{0.5,0.5,0.5}}%
\else
\def\colorrgb#1{\color{black}}%
\def\colorgray#1{\color[gray]{#1}}%
\expandafter\def\csname LTw\endcsname{\color{white}}%
\expandafter\def\csname LTb\endcsname{\color{black}}%
\expandafter\def\csname LTa\endcsname{\color{black}}%
\expandafter\def\csname LT0\endcsname{\color{black}}%
\expandafter\def\csname LT1\endcsname{\color{black}}%
\expandafter\def\csname LT2\endcsname{\color{black}}%
\expandafter\def\csname LT3\endcsname{\color{black}}%
\expandafter\def\csname LT4\endcsname{\color{black}}%
\expandafter\def\csname LT5\endcsname{\color{black}}%
\expandafter\def\csname LT6\endcsname{\color{black}}%
\expandafter\def\csname LT7\endcsname{\color{black}}%
\expandafter\def\csname LT8\endcsname{\color{black}}%
\fi
\fi
\setlength{\unitlength}{0.0500bp}%
\begin{picture}(7200.00,5040.00)%
\gplgaddtomacro\gplbacktext{%
	\csname LTb\endcsname%
	\put(1078,704){\makebox(0,0)[r]{\strut{} 0.6}}%
	\csname LTb\endcsname%
	\put(1078,1213){\makebox(0,0)[r]{\strut{} 0.65}}%
	\csname LTb\endcsname%
	\put(1078,1722){\makebox(0,0)[r]{\strut{} 0.7}}%
	\csname LTb\endcsname%
	\put(1078,2231){\makebox(0,0)[r]{\strut{} 0.75}}%
	\csname LTb\endcsname%
	\put(1078,2740){\makebox(0,0)[r]{\strut{} 0.8}}%
	\csname LTb\endcsname%
	\put(1078,3248){\makebox(0,0)[r]{\strut{} 0.85}}%
	\csname LTb\endcsname%
	\put(1078,3757){\makebox(0,0)[r]{\strut{} 0.9}}%
	\csname LTb\endcsname%
	\put(1078,4266){\makebox(0,0)[r]{\strut{} 0.95}}%
	\csname LTb\endcsname%
	\put(1078,4775){\makebox(0,0)[r]{\strut{} 1}}%
	\csname LTb\endcsname%
	\put(1210,484){\makebox(0,0){\strut{} 0.01}}%
	\csname LTb\endcsname%
	\put(4006,484){\makebox(0,0){\strut{} 0.1}}%
	\csname LTb\endcsname%
	\put(6803,484){\makebox(0,0){\strut{} 1}}%
	\put(176,2739){\rotatebox{-270}{\makebox(0,0){\strut{}Rate of classifying Benign as Benign}}}%
	\put(4006,154){\makebox(0,0){\strut{}Rate of classifying Porn as Benign}}%
}%
\gplgaddtomacro\gplfronttext{%
	\csname LTb\endcsname%
	\put(5816,1757){\makebox(0,0)[r]{\strut{}ANet-Fulltrain}}%
	\csname LTb\endcsname%
	\put(5816,1537){\makebox(0,0)[r]{\strut{}GNet-Fulltrain}}%
	\csname LTb\endcsname%
	\put(5816,1317){\makebox(0,0)[r]{\strut{}ANet}}%
	\csname LTb\endcsname%
	\put(5816,1097){\makebox(0,0)[r]{\strut{}GNet}}%
	\csname LTb\endcsname%
	\put(5816,877){\makebox(0,0)[r]{\strut{}AGNet}}%
}%
\gplbacktext
\put(0,0){\includegraphics{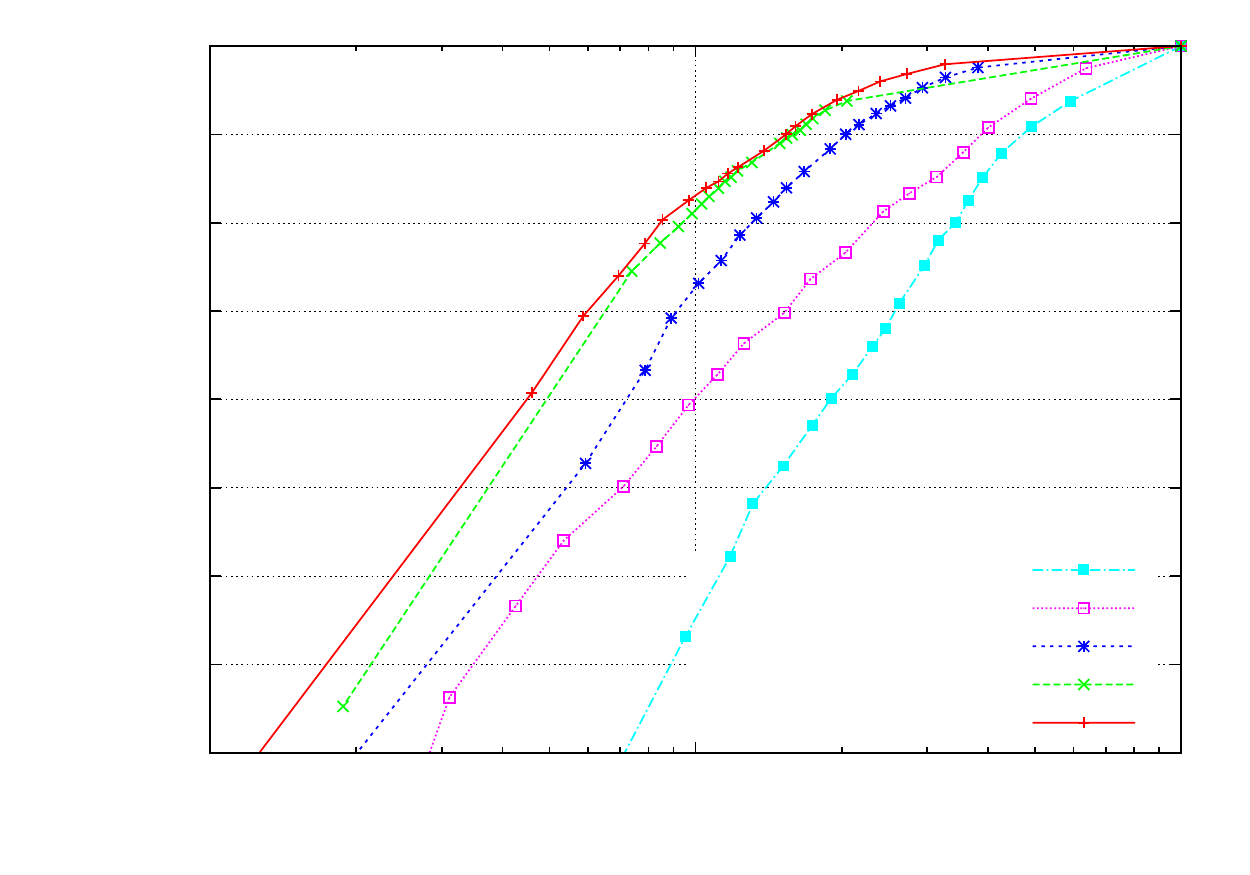}}%
\gplfronttext
\end{picture}%
\endgroup

	\caption{ConvNet Benign/Porn image classifier ROC.}
	\label{fig:imageroc}
\end{figure}

We can see in figure \ref{fig:imageroc}, that both ANet and GNet (which were fine-tune trained as described earlier in the previous section) produced higher ROC than their fully trained counterparts. For instance when the rate of classifying porn frames by mistake as benign was 0.1, i.e., 10\%, the rate of classifying benign frames correctly stood at 0.68 and 0.87 respectively for ANet fully trained and ANet fine tuned. Similarly, rates for GNet fully trained and fine tuned were 0.80 and 0.91 respectively. The ROC curves also show that AGNet is more accurate than GNet, especially when the rate of false porn classification is very low or very high.

\subsection{Experiment: testing videos}
In this experiment, we have considered the majority voting of all frames, that belong to the same video sequence, in the decision as described earlier in the AGNet section. We recorded the average correct classification rate of all 5 folds and the standard deviation. The reported correct classification rate is the average of both `correctly classifying benign as benign' and `correctly classifying porn as porn'.  We included in table \ref{table:NPDI video results} also the best accuracy numbers reported in the recent literature as a baseline for comparison.

As expected, GNet is more accurate than ANet as it has deeper architecture. Additionally, AGNet (with simple equal weighted average) is slightly more accurate than GNet. Note also that AGNet produced smaller variance than either ANet or GNet. For the record, we also reported the results of AGbNet, where the score `fusion' is the larger of ANet and GNet score instead of the simple average.

It was interesting to see the all ConvNet based classifiers outperformed the BossaNova state of the art methods \cite{Avila2013453} \cite{Caetano2014} in the average accuracy but with slightly larger standard deviation. This might be due to the random initialization of the neural networks weights during training. We believe that the variance would go lower if we were to add more networks to the ensemble and fine tune some training parameters.

\begin{table}
	\caption{Videos classification accuracy on NPDI benchmark dataset}
	\centering
	\begin{tabular}{cc}
		\hline \hline
		Approach & Accuracy (\%) \\ 
		\hline
		BossaNova (HueSIFT) \cite{Avila2013453} & 89.5$\pm$1 \\ 
		BossaNova VD (BinBoost16) \cite{Caetano2014} & 90.9$\pm$1 \\ 
		Proposed ANet & 92.01$\pm$3 \\
		Proposed GNet & 93.7$\pm$3 \\
		Proposed AGNet & \textbf{93.8$\pm$2} \\
		Proposed AGbNet & \textbf{94.1$\pm$2} \\
		
		\hline 
	\end{tabular}
	\label{table:NPDI video results}
\end{table}

\section{Conclusions}
We proposed applying convolutional neural networks to automatically classify pornographic images and videos. We showed that our proposed fully automated solution outperformed the accuracy of hand-crafted feature descriptors solutions. We are continuing our research to find an even better network architecture for this problem. Nevertheless, all the successful applications so far rely on supervised training methods. We expect a new wave of deep learning networks would emerge by combining supervised and unsupervised methods where a network can learn from its mistakes while in actual deployment.
We believe further research can also be directed toward allowing machines to consider the context and overall rhetorical meaning of a video clip while relating them to the images involved.

\end{document}